\documentclass{article}
\usepackage{spconf,amsmath,epsfig}
\usepackage[noadjust]{cite}
\usepackage{amssymb}
\usepackage{amsfonts}
\usepackage{mathtools}
\usepackage{bbm}
\usepackage{pifont}

\usepackage{upgreek}
\usepackage{float}
\usepackage{subcaption}
\usepackage{multicol}
\usepackage{multirow}
\usepackage{verbatim}
\usepackage{soul}

\usepackage{amsmath,amssymb,amsfonts}
\usepackage[linesnumbered,lined,boxed,commentsnumbered,ruled]{algorithm2e}
\usepackage[usenames,dvipsnames]{xcolor} 

\usepackage{adjustbox}
\usepackage{makecell}
\usepackage{comment}

\DeclareMathAlphabet\mathbfcal{OMS}{cmsy}{b}{n}
\title{Transformer-based Federated Learning for Multi-Label Remote Sensing Image Classification}
\name{Bar{ı}\c{s} B\"{u}y\"{u}kta\c{s}{\normalfont\textsuperscript{1,2}}, Kenneth Weitzel {\normalfont\textsuperscript{2}}, Sebastian V\"{o}lkers{\normalfont\textsuperscript{2}}, Felix Zailskas{\normalfont\textsuperscript{2}}, Beg\"{u}m Demir{\normalfont\textsuperscript{1,2}}}
\address{
\textsuperscript{1}BIFOLD - Berlin Institute for the Foundations of Learning and Data, Germany\\
\textsuperscript{2}Faculty of Electrical Engineering and Computer Science, Technische Universit\"at Berlin, Germany}

\begin{document}
%
\maketitle

\begin{abstract}
Federated learning (FL) aims to collaboratively learn deep learning model parameters from decentralized data archives (i.e., clients) without accessing training data on clients. However, the training data across clients might be not independent and identically distributed (non-IID), which may result in difficulty in achieving optimal model convergence. In this work, we investigate the capability of state-of-the-art transformer architectures (which are MLP-Mixer, ConvMixer, PoolFormer) to address the challenges related to non-IID training data across various clients in the context of FL for multi-label classification (MLC) problems in remote sensing (RS). The considered transformer architectures are compared among themselves and with the ResNet-50 architecture in terms of their: 1) robustness to training data heterogeneity; 2) local training complexity; and 3) aggregation complexity under different non-IID levels. The experimental results obtained on the BigEarthNet-S2 benchmark archive demonstrate that the considered architectures increase the generalization ability with the cost of higher local training and aggregation complexities. On the basis of our analysis, some guidelines are derived for a proper selection of transformer architecture in the context of FL for RS MLC. The code of this work is publicly available at https://git.tu-berlin.de/rsim/FL-Transformer.

\end{abstract}
\section{Introduction}
\label{sec:intro}

Recent advances in satellite technologies lead to a significant increase in the amount of remote sensing (RS) data that is distributed across decentralized image archives (i.e., clients). The development of accurate multi-label classification (MLC) methods in massive archives of RS images is a growing research interest in RS. Deep learning (DL) methods have achieved remarkable success in MLC of RS images. However, employing DL typically necessitates full access to training data for learning model parameters \cite{burgert2022effects, sumbul2023generative, mollenbrok2023deep}. In some clients, training data may not be publicly accessible due to commercial interests, privacy concerns, and legal regulations \cite{papernot2016semi}. 
To address this problem, federated learning (FL) that allows to jointly train a DL model without any access to the data on the clients (i.e., decentralized image archives) and to find the optimal model parameters of a global model on a central server can be used. 
 
It is worth noting that the training data in different clients might be not independent and identically distributed (non-IID). The presence of non-IID data in FL can reduce the overall MLC performance as it affects the convergence of the global model \cite{li2021model}. Although there are many studies that aim to mitigate the effects of non-IID in computer vision \cite{li2021model,gao2022feddc,mendieta2022local,ma2022layer,Li:2023,qin2023reliable}, it is seldom studied in RS \cite{buyuktacs2023federated,zhang2022prototype,zhang2023federated}. As an example, in \cite{zhang2022prototype}, an FL algorithm is introduced, aiming to train multiple global models using hierarchical clustering, when RS images are non-IID among clients. There are also other studies in RS related to FL but addressing different problems \cite{buyuktacs2023learning,zhang2023local,chen2024free}. As an example, in \cite{buyuktacs2023learning}, an FL framework that learns DL model parameters from decentralized and unshared multi-modal data archives is introduced. In \cite{zhang2023local}, local differential privacy federated learning algorithm is proposed to collaboratively train DL models while secured by local differential privacy. Most of the methods that address training data heterogeneity aim to improve either standard empirical risk minimization of local training or standard parameter aggregation of model averaging \cite{buyuktacs2023federated}. In \cite{qu2022rethinking}, it is shown that Vision Transformers (ViTs) \cite{dosovitskiy2020image} and Swin Transformers \cite{liu2021swin} are beneficial to alleviate the effects of training data heterogeneity in FL models. In \cite{qu2022rethinking}, non-IID data partitions are simulated by only considering label distribution skew in the context of single-label classification problems. In MLC problems, different clients might have varying combinations of labels associated with their data, potentially leading to even more skewed label distributions across clients. In addition to that, there might be other factors that cause non-IID data partitions. In detail, various clients may have different amounts of training data (i.e., quantity skew) \cite{kairouz2021advances}. Also, the images that belong to the same class might exhibit different data distributions among different clients (i.e., concept drift) \cite{li2021fedbn}, which may reduce the generalization capability of DL models.

In this paper, we focus our attention on the state-of-the-art transformer architectures (MLP-Mixer \cite{tolstikhin2021mlp}, ConvMixer \cite{trockman2022patches}, PoolFormer \cite{yu2022metaformer}) and present a study to analyze and compare them both theoretically and experimentally in the framework of FL for RS MLC. In particular, we assess their capability to address training data heterogeneity when quantity skew, label distribution skew, and concept drift exist across decentralized RS image archives.

\section{Methodology}

\subsection{Problem Statement}
 Let $K$ be the total number of clients and $C^i$, $1\leq i\leq K$ be the $i$th client. In detail, $C^i$ holds the corresponding training set $D_i = \{(\boldsymbol{x}_{z}^i, \boldsymbol{y}_{z}^i)\}_{z=1}^{M^i}$, where $M^i$ is the number of images, $\boldsymbol{x}_{z}^i$ is the $z$th RS image of the $i$th client and $\boldsymbol{y}_{z}^i$ is the corresponding class label. $\boldsymbol{y}_{z}^i =[{y}_{z,1}^i,..., {y}_{z,P}^i] \in \{ 0, 1 \}^P$ is a multi-label vector of $\boldsymbol{x}_{z}^i$, stating the presence of $P$ unique classes (i.e., ${y}_{z,p}^i =1$ indicates that $p$-th class is present and ${y}_{z,p}^i =0$ indicates that $p$-th class is not present in $\boldsymbol{x}_{z}^i$). There is at least one class label associated with each image. The DL model $\phi^i$ is trained using $D_i$ for each $C^i$. To this end, we use binary cross-entropy (BCE) loss, which is given for a single sample as follows:
\begin{equation}
\begin{split}
    \mathcal{L}_{BCE}(\phi^i(\boldsymbol{x}_{z}^i), \boldsymbol{y}_{z}^i) = & \\
    - \sum^P_{p=1} {y}_{z,p}^i \log( & r_{z,p}^i) +  (1-{y}_{z,p}^i \log(1-r_{z,p}^i)
\end{split}
\end{equation}
where $\phi^i(\boldsymbol{x}_{z}^i) = [r_{z,1}^i, ...,r_{z,P}^i]$ is the class probability vector obtained by the DL model. In this paper, we assume that the data in clients is not shared and FL is used to learn global model parameters $w$ over the whole training set $M = \bigcup_ {i \in \{ 1, 2,..., K \} } M^i$ while minimizing the following global objective function as follows: 

\begin{equation}
\arg \min_{w} L(w) =  \sum_{i=1}^{K} \frac{ M^i}{\lvert M \rvert} L_{i}(w),
\label{minimization}
\end{equation}
where $L_{i}$ is the empirical loss of $C^i$. The parameter update of the algorithm is as follows:

\begin{equation}
w = \sum_{i=1}^{K} \alpha_i  w^{i},
\label{fedavg}
\end{equation}
where $\alpha_i$ is a hyperparameter to adjust the importance of the local parameters in each client for aggregation, $w^{i}$ is the local model parameters of $C^i$. 

 In this paper, we aim to study the effectiveness of transformer architectures in terms of their: 1) robustness to training data heterogeneity; 2) local training complexity; and 3) aggregation complexity in FL settings for RS MLC problems under non-IID data. In particular, we focus on three architectures: 1) MLP-Mixer; 2) ConvMixer; and 3) PoolFormer. As FL algorithms, we select federated averaging (FedAvg) \cite{mcmahan2017communication} (which does not explicitly address training data heterogeneity) and model-contrastive federated learning (denoted as MOON) \cite{li2021model} (which is  known for its strong performance in the case of data heterogeneity among clients).

\subsection{Selected Federated Learning Algorithms}

As one of the first studies, the FedAvg algorithm is introduced in \cite{mcmahan2017communication} to train the local models using local data on clients and the central server aggregates the updated model parameters through iterative model averaging. There are four steps of iterative model averaging: 1) the server sends the global model to clients; 2) the clients update the model parameters; 3) the clients send the model parameters to central server; and 4) the central server aggregates the model parameters.

In \cite{li2021model}, the MOON algorithm is introduced to address the challenges of training data heterogeneity by adding the proximal term to the local objective function of each client. The aim of this algorithm is to limit the size of local updates, leading to more robust performance when the training data is non-IID. The proximal term is calculated by increasing the similarity between the image features obtained from the global model and the local model, and reducing the similarity of features obtained between current and previous local models. The proximal term of the MOON algorithm is defined as follows:
\begin{equation}
l_{con}=-\frac{e^ { S( z_{t+1}, z ) / \tau  } }  {  e^ { S( z_{t+1}, z ) / \tau  } + e^ { S( z_{t}, z ) / \tau  }}
\label{moon}
\end{equation}
where $S(.,.)$ measures the cosine similarity, $\tau$ is a temperature parameter. $z_{t+1}$, $z_{t}$, $z$ are the representation vectors from the current local model $w_{t+1}^i$, previous local model $w_{t}^i$, global model $w$, respectively.

\subsection{Selected Transformer Architectures}

\subsubsection{MLP-Mixer}

The MLP-Mixer \cite{tolstikhin2021mlp} is introduced as an alternative computer vision model to CNNs and the Vision Transformers (ViT) \cite{dosovitskiy2020image} to reduce the computational time. Similar to the ViT, the MLP-Mixer employs the concept of channel mixing and token mixing to enable communication across different channels as well as spatial locations. The token mixing and channel mixing operations allow the local models to learn diverse representations of the input data. This capability enables global model to adapt to variations in data distributions across various clients. This can help global model to be more robust to training data heterogeneity. Unlike many other transformer architectures, it does not use any self-attention or convolutions but rather only utilizes multi-layer perceptrons (MLPs) that are applied to patches of an input image. The absence of convolutional layers and self-attention mechanisms reduce the computational costs required to train a DL model per client (i.e., local training complexity), as there is no need to perform convolution operations or maintain convolutional kernels. Since MLP-Mixer leverages a simple parameter structure, it results in using fewer trainable model parameters. This reduces the cost of aggregating model parameters at the central server (i.e., aggregation complexity).

\subsubsection{ConvMixer}
The ConvMixer architecture \cite{trockman2022patches} builds upon the MLP-Mixer by including channel and token mixing mechanisms to process channel and spatial features. The ConvMixer replaces all MLP operations in MLP-Mixer with convolutional layers. The image is initially embedded in a patchwise manner via a convolutional layer. Then, the patch embedding is passed to a depthwise convolutional layer. The following pointwise convolutional layer performs $1x1$ convolution to achieve channel mixing. Since ConvMixer does not use any attention mechanism, it may struggle to focus on important features or relationships within the data. This can reduce the generalization capability of global model in FL setups with heterogeneous data distributions. The ConvMixer architecture typically requires a large number of convolutional filters across its layers to capture spatial hierarchies effectively. The increased number of filters may lead to high local training complexity. Furthermore, the architectural choices regarding to the number of layers and filters can notably increase the aggregation complexity, given that each filter encompasses a set of trainable parameters.

\subsubsection{PoolFormer}

The PoolFormer architecture \cite{yu2022metaformer} replaces the attention-based token mixer module by the simple average pooling operation as a token mixer. The average pooling operation provides a summary representation of the most important features within each pooling window. As clients obtain smoother local model parameters, the global model becomes more robust to the limitations of training data heterogeneity. Since the self-attention layers introduce quadratic runtime complexity, the absence of self-attention reduces local training complexity. The average pooling layers do not introduce additional parameters to be learned. Therefore, the aggregation complexity of PoolFormer is lower than those of MLP-Mixer and ConvMixer.

\section{Experimental Results}

The experiments were conducted on the BigEarthNet-S2 benchmark archive \cite{sumbul2021bigearthnet}. In detail, we used a subset of BigEarthNet-S2 that includes 52,906 Sentinel-2 images acquired over Austria, Belgium, Finland, Ireland, Lithuania, Serbia and Switzerland in summer. Each image is annotated with multi-labels provided by the CORINE Land Cover Map database. We used the 19 class nomenclature introduced in \cite{sumbul2021bigearthnet}. In the experiments, we used the train and test split proposed in \cite{sumbul2021bigearthnet}. To assess the architectures on different non-IID levels, we designed two different data partitions (i.e., scenarios) on the official train split as follows:
\begin{itemize}
       \item \textit{Decentralization Scenario 1 (DS1)}: The training images were randomly distributed to different clients.
       \item \textit{Decentralization Scenario 2 (DS2)}: The training images were distributed in a way that each client holds the images associated with only one country.
\end{itemize}
Since the images are randomly distributed to the clients in DS1, the level of non-IID in DS1 is lower than that of DS2. We have compared the results of MLP-Mixer \cite{tolstikhin2021mlp}, ConvMixer \cite{trockman2022patches} and PoolFormer \cite{yu2022metaformer} architectures with those of the ResNet-50 in the context of FedAvg \cite{mcmahan2017communication} and MOON \cite{li2021model} algorithms for RS MLC. The models were trained for 30 communication rounds and three epochs using Adam optimizer with the learning rate of 0.001, mini-batch size of 128. We compare the performance of architectures in terms of MLC accuracy (in $F_1$-Score), local training complexity, and aggregation complexity. Table \ref{tab:DS_1} shows the $F_1$-scores (\%) under both scenarios. One can see from the table that the results under DS1 are higher than those under DS2. This is due to the fact that the level of training data heterogeneity of DS1 is lower than that of DS2. According to the results obtained using FedAvg algorithm, the highest accuracy was obtained by ResNet-50, which is 2.77\% higher than that obtained by the worst performing architecture (i.e., ConvMixer) under DS1. In detail, the results obtained with MLP-Mixer and PoolFormer architectures are similar to each other and higher than those of ConvMixer by almost 1.5\%. Unlike the results obtained under DS1, the lowest accuracy was obtained with ResNet-50 under DS2. This indicates that the transformer architectures more effectively handle the challenges of training data heterogeneity and increase the accuracy of the global model. The highest accuracy was achieved with PoolFormer under DS2, which is 12.53\% higher than the results obtained with ResNet-50. According to the results obtained using MOON algorithm, the $F_1$-scores obtained under DS2 are higher than those obtained with FedAvg due to the capability of the MOON algorithm to address the limitations of training data heterogeneity. However, this increase is less prominent with transformer architectures. As an example, although MOON algorithm outperforms FedAvg algorithm by 7.51\% higher $F_1$-score with Resnet-50, MOON algorithm does not significantly increase the $F_1$-score with PoolFormer. This shows that the transformer architectures already alleviate the effects of training data heterogeneity. Therefore, it is feasible to achieve high $F_1$-score without employing an FL method that specifically tackles non-IID data.

\begin{table}[t]
\renewcommand{\arraystretch}{1.1}
\setlength\tabcolsep{13pt}
\caption{$F_1$-scores (\%) obtained by FedAvg and MOON algorithms with different architectures under DS1 and DS2.}
\label{tab:DS_1}
\centering
\begin{tabular}{@{}llccc@{}}
\hline
Algorithms & Architectures & DS1 & DS2  \\ \hline
\multirow{4}{*}{\thead[l]{FedAvg \cite{mcmahan2017communication}}} & ResNet-50 & 75.24  & 47.32   \\ 
& MLP-Mixer \cite{tolstikhin2021mlp} & 73.87   & 59.79      \\ 
& ConvMixer \cite{trockman2022patches}   & 72.47    & 53.66    \\ 
& PoolFormer \cite{yu2022metaformer}   & 74.09    & 59.85  \\ \hline
\multirow{4}{*}{\thead[l]{MOON \cite{li2021model}}}  
& ResNet-50  & 72.82    & 54.83  \\
& MLP-Mixer \cite{tolstikhin2021mlp}  & 74.83    &60.05   \\ 
& ConvMixer \cite{trockman2022patches}  & 73.28    & 58.76   \\ 
 & PoolFormer \cite{yu2022metaformer}   & 74.47    & 60.26    \\ \hline
\end{tabular}
\end{table}

We also assess the effectiveness of the architectures in terms of local training complexity and aggregation complexity. Table \ref{tab:DS_2} shows the computational times (in seconds) to complete a single communication round of local training. By analyzing the table, one can see that the computational time of MOON algorithm is higher than that of FedAvg algorithm for all selected architectures. This is because of the fact that the MOON algorithm requires extra forward passes to extract image feature vectors. One can observe from the table that the computational time required with ResNet-50 is the lowest. The local training complexity of MLP-Mixer and PoolFormer are similar and lower than that of ConvMixer. Table \ref{tab:DS_2} shows the number of model parameters shared by clients. According to the results, PoolFormer is the most efficient architecture among all the selected architectures in terms of aggregation complexity. As an example, the number of parameters of a model that is trained by PoolFormer is less than that of other selected architectures.

\begin{table}[t]
\renewcommand{\arraystretch}{1.1}
\setlength\tabcolsep{7.5pt}
\caption{Computational times (in seconds) and number of parameters shared by each client with different architectures obtained by FedAvg and MOON algorithms.}
\label{tab:DS_2}
\centering
\begin{tabular}{@{}llccc@{}}
\hline
Algorithms & Architectures & \thead[c]{Computational \\ Time}  &\thead[c]{ Number of \\ Shared \\ Parameters}   \\ \hline
\multirow{4}{*}{\thead[l]{FedAvg \cite{mcmahan2017communication}}} & ResNet-50 & 767  & 23.60M   \\ 
& MLP-Mixer \cite{tolstikhin2021mlp} & 1092    & 18.34M      \\ 
& ConvMixer \cite{trockman2022patches}   & 2432    & 20.62M    \\ 
& PoolFormer \cite{yu2022metaformer}   & 1231    & 11.24M  \\ \hline
\multirow{4}{*}{\thead[l]{MOON \cite{li2021model}}}  
& ResNet-50  & 972    & 23.60M  \\
& MLP-Mixer \cite{tolstikhin2021mlp}  & 1323    & 18.34M  \\ 
& ConvMixer \cite{trockman2022patches}  & 2683    & 20.62M    \\ 
 & PoolFormer \cite{yu2022metaformer}   & 1447    & 11.24M    \\ \hline
\end{tabular}
\end{table}

\section{Conclusion}

This paper analyzes and compares different transformer-based architectures (MLP-Mixer, ConvMixer, PoolFormer) under different non-IID levels in the framework of FL for MLC problems in RS. The selected architectures have been compared in term of their: 1) robustness to training data heterogeneity; 2) local training complexity; and 3) aggregation complexity (see Table \ref{comp}). Through an experimental comparison of these architectures, we have derived a guideline for selecting appropriate architectures in the context of FL for RS MLC problems as follows:
\begin{enumerate}
 \item The architectures achieve similar accuracies when the level of training data heterogeneity is low. ResNet-50 has the lowest local training complexity. Therefore, although the aggregation complexity of ResNet-50 is the highest among transformer architectures, it can be selected if training data heterogeneity is low.
 \item The accuracy of ResNet-50 drops significantly when the level of training data heterogeneity increases. MLP-Mixer and PoolFormer achieve the highest accuracies among the selected architectures when the level of training data heterogeneity is high. Since the aggregation complexity of PoolFormer is lower than that of MLP-Mixer, it can be selected if the training data is highly non-IID.
 \item MOON outperforms FedAvg by a large margin when the level of training data heterogeneity is high with ResNet-50. However, the accuracy obtained with MOON is very close to that of FedAvg when the transformer architectures are used. Therefore, the FedAvg algorithm can be used with transformers instead of complex algorithms that alleviate the effects of training data heterogeneity. 
\end{enumerate}

As a final remark, we would like to note that, although MLC has been selected as the classification task to assess different transfomer-based architectures for FL in this paper, our study can be extended to the other learning tasks such as semantic segmentation. As a future work, we plan to conduct research in this direction.

\begin{table}[t]
\renewcommand{\arraystretch}{1.1}
\setlength\tabcolsep{5.3pt}
\caption{Comparison of the selected architectures. Three marks "H" (High), "M" (Medium), or “L” (Low) are given for the considered criteria.}
\centering
\begin{tabular}{@{}llcccc@{}}
\hline
Architectures & & \thead[c]{Accuracy} & \thead[c]{Local Training\\Complexity} & \thead[c]{Aggregation\\Complexity}  \\ \hline
ResNet-50  &  &  {L}  &  {L}   &  {H}            \\ \hline
MLP-Mixer \cite{tolstikhin2021mlp}  &  &  {H}  &  {M}   &  {H}            \\ \hline
ConvMixer \cite{trockman2022patches}  &  &  {M}  &  {H}   &  {H}            \\ \hline
PoolFormer \cite{yu2022metaformer}  &  &  {H}  &  {M}   &  {L}            \\ \hline
\end{tabular}
\label{comp}
\end{table}

\section{ACKNOWLEDGEMENTS}
This work is supported by the European Research Council (ERC) through the ERC-2017-STG BigEarth Project under Grant 759764.

\bibliographystyle{IEEEtran}
{\small
\bibliography{strings,refs}}
\end{document}